\documentclass[11pt]{article}

\usepackage{acl}

\usepackage{times}
\usepackage{latexsym}

\usepackage{pgfplots}
\pgfplotsset{compat=1.18}
\usepackage[T1]{fontenc}

\usepackage[utf8]{inputenc}

\usepackage{microtype}

\usepackage{inconsolata}

\usepackage{graphicx}
\graphicspath{{images/}}

\usepackage{tikz}
\usetikzlibrary{arrows.meta,positioning,shapes.geometric,calc}
\usepackage{adjustbox}

\usepackage{mystyle}

\title{OUNLP at TSAR 2025 Shared Task: Multi-Round \\ Text Simplifier via Code Generation}

\author{Cuong Huynh \\
  School of Computer Science \\
  University of Oklahoma\\
  \texttt{cuong@ou.edu} \\\And
  Jie Cao \\
  School of Computer Science \\
  University of Oklahoma\\
  \texttt{jie.cao@ou.edu} \\}

\begin{document}
\maketitle

\begin{abstract}
This paper describes the OUNLP system submitted to the TSAR-2025 Shared Task~\cite{alva-manchego-etal-2025-findings}, designed for readability-controlled text simplification using LLM-prompting-based generation. Based on the analysis of prompt-based text simplification methods, we discovered an interesting finding that text simplification performance is highly related to the gap between the source CEFR~\cite{arase2022cefr} level and the target CEFR level. Inspired by this finding, we propose two multi-round
simplification methods and generate them via GPT-4o: rule-based simplification (MRS-Rule) and jointly rule-based LLM simplification (MRS-Joint). Our submitted systems ranked 7 out of 20 teams. Later improvements with MRS-Joint show that taking the LLM simplified candidates as the starting point could further boost the multi-round simplification performance~\footnote{\url{https://github.com/ounlp/Multi-Round-Text-Simplifier}}.
\end{abstract}

\section{Introduction}
Complex text makes it difficult for language learners and people with limited literacy to read. Text simplification improves learning, accessibility, and information sharing with a wider audience. With the advent of deep learning and large language models (LLMs), simplification performance has improved significantly, supported by the release of important datasets~\cite{imperial2025universalcefr}. Modern approaches have explored zero-shot prompting ~\cite{chi-etal-2023-learning,barayan-etal-2025-analysing,farajidizaji-etal-2024-possible}, instruction tuning~\cite{imperial-tayyar-madabushi-2023-flesch}, and related strategies.

From our baseline analysis of trial data, we observed that a larger gap between the CEFR level of the original sentence and the target level (\textbf{CEFR-Gap}) substantially increases the likelihood of simplification failure. This finding highlights the importance of addressing complexity not in a single step but through a structured, iterative process. Building on this insight, we introduced two novel models generated by GPT-4o for multi-round text simplification: MRS-Rule, a rule-based framework that progressively adjusts sentence structures and vocabulary, and MRS-Joint, which integrates rules with prompting techniques to leverage the strengths of both symbolic and generative approaches.

The primary contribution of this work is to show that multi-round small rule-based simplification are more effective at handling large CEFR gaps than conventional single-step approaches. Our proposed MRS-Joint method outperforms the MRS-Rule and baseline models, as validated through extensive experiments and qualitative analyzes. Additionally, we explore the potential of automatic code generation for text simplification, although further refinement remains necessary.

\section{Task Setup}
The goal of the shared task is to simplify a given source text into a target text with the desired CEFR proficiency level (A1$<$A2$<$B1$<$B2$<$C1$<$C2). For the datasets, we use the same trial~(40 examples) and test~(200 examples) data sets provided by the TSAR workshop to build and evaluate our methods. For the evaluation metrics, we follow the same metrics from the official TSAR-2025 shared-task metrics, which covers both readability-level control~(CEFR Compliance, we focus on~\textbf{RMSE}, the distance between predicted and target CEFR levels, the lower the better) and the preservation of meaning by evaluating semantic fidelity between the simplified sentence and the original sentence, or the simplified sentence and a human-written reference via \textbf{MeaningBERT} \citep{beauchemin2023meaningbert}, denoted as \textbf{MB-Orig} and \textbf{MB-Ref} respectively~\footnote{Please refer to the shared task paper~\cite{alva-manchego-etal-2025-findings} for more details of other metrics such as BERTScore~\cite{zhang2019bertscore} etc.}.

\section{Motivation for Multi-Round}
\label{sec:pg1}
In this section, we present our baseline model, the \Naive~Prompt model, and show that simplification becomes increasingly challenging as the gap between the source and target levels widens.

\subsection{Baseline: \Naive~Prompt-based (Run 1)}
\label{ssec:baseline}
 We use GPT-4o to generate the code first~(denoted as Baseline or "Program 1"), which will call the OpenAI APIs~(GPT-4o-mini) with the following prompt from~\cite{barayan-etal-2025-analysing}. This generated our Run-1 submission of the test data. Please refer to Appendix~\ref{ssec:detail-program1} for more details. 

\begin{tcolorbox}[promptbox, title={Baseline Prompt}]
Please simplify the following Complex Sentence to make it easier to read and understand by \{CEFR-LEVEL\} CEFR level English learners. \{CEFR-LEVEL\} level English learner \{CEFR-Description\}. To simplify, you may replace difficult words with simpler ones, elaborate, or remove them when possible. You may also break down a lengthy sentence into shorter, clear sentences. Ensure the revised sentence is grammatically correct, fluent, and maintains the core message of the original without changing its meaning. Complex Sentence: \{Source\} Simplified Sentence:
\end{tcolorbox}

\paragraph{CEFR Level Prediction}
Since the trial data only gives the CEFR level for target text, not for the source text and the simplified texts, we estimate a text proficiency level using three ModernBERT classifiers with the voting mechanism~\footnote{
AbdullahBarayan/ModernBERT-base-doc\_en-Cefr, ModernBERT-base-doc\_sent\_en-Cefr, and
ModernBERT-base-reference\_AllLang2-Cefr2}. Each model independently predicts a \textsc{CEFR} label (A1-C2) with a confidence score. We combine predictions via majority voting: the label with the most votes is selected. Ties are broken by the largest sum of confidences, then by the highest single-model confidence; if still tied, we prefer the simpler (lower) level to remain conservative. The resulting CEFR level also determines whether a simplification is
still needed for a text.
 \paragraph{CEFR-Gap} We assign an integral value from 0 to 5 for each CEFR level according to the order of~(A1$<$A2$<$B1$<$B2$<$C1$<$C2). The CEFR gap for each example is defined as the numerical difference between the source level and the targe level~(e.g., the CEFR gap between C1 and A2 is 4-1=3).  We run the generated program on the 40 trial examples in the trial data, and then study the performance of the baseline models for each group of examples with the same CEFR Gap as Table~\ref{tab:baseline-gap}. We found that \textbf{RMSE} rises from 0.624 with a one-level gap,  to 1.027 with two levels,  and then goes further to 1.581 with three levels, indicating that larger downward steps are harder to control. Meaning preservation also weakens: \textbf{MB-Orig} declines from 0.859 to 0.841 and then to 0.761, while \textbf{MB-Ref} falls from 0.832 to 0.758 and stays near 0.762 for the widest gap, although that last figure is based on only four samples. These patterns reveal a trade-off: stronger simplification with larger \textbf{CEFR-Gap} makes it more difficult to match the target level and to keep the original meaning intact. In short, bigger CEFR gaps demand more radical linguistic changes, which inevitably reduce both level accuracy and semantic fidelity.

\begin{table}[!tp]
\centering
\small
\setlength{\tabcolsep}{6pt}
\begin{tabular}{c|c|c|c}
\toprule
\textbf{CEFR-Gap} & \textbf{RMSE} & \textbf{MB-Orig} & \textbf{MB-Ref} \\
\midrule
1 (18) & 0.624& 0.859& 0.832 \\
2 (18) &  1.027& 0.841&  0.758\\
3 (4) &  1.581& 0.761&  0.762\\
\bottomrule
\end{tabular}
\caption{CEFR-Gap Analysis on CEFR accuracy~(RMSE) and meaning preservation. The bracket shows the total number of examples we found in the trial data with that CEFR gap. It shows the larger the gap, the higher the RMSE, the lower the other MB scores.}
\label{tab:baseline-gap}
\vspace{-2em}
\end{table}

\section{Proposed Multi-Round Methods}
\label{sec:methods}
Based on the findings in ~\S\ref{sec:pg1}, smaller gap between the source and the target CEFR level will be relatively easy to simplify. Hence, we propose to simplify texts with multiple rounds by taking previous simplification results as inputs with two multi-round methods: rule-based simplification~(MRS-Rule~\S\ref{ssec:pg2}) and jointly rule-based and LLM Prompting~(MRS-Joint~\S\ref{ssec:pg3}). For each program, we first demonstrate the prompts and operations to generate and fix, and then briefly analyze the detailed workflow of the generated program. The orange box shows the operations and prompts we used to generate the MRS-Rule Code, while the blue box at the bottom shows the further steps we used to fix the generated code to make it work. 

\subsection{MRS-Rule: Rule-based~(Run 2)}
\label{ssec:pg2}
The generated code~(see more details in Appendix \S\ref{ssec:detail-program2}) for MRS-Rule does not call any large language model API for simplification, but only rule-based rewriting combined with automatic CEFR level verification and semantic checks.
\begin{tcolorbox}[promptbox,enhanced,title={Prompts for Generating MRS-Rule Code},breakable=false]
<Operations:> Upload the Program 1~(Baseline) file into GPT-4o; Upload 2 images (one image is for the three models, and the other is the method to predict the level in the evaluation).

\textbf{Prompt 2.1} I want you to write me a program that simplifies the original sentence. In the program, the first step is to simplify the original sentence. The next step is to identify the CEFR\_Level using three models and the method in the image. If the generated Cefr level does not match the target level, it will call the simplify sentence method to simplify that simplified sentence until it reaches the target level. If the generated level matches the target level, it will be written in the output file. Write down the program based on the file (program used only naive prompt) and 2 images (1 is three model, and the other is the method to predict the level) I give you.

\medskip
\textbf{Prompt 2.2} The program keeps the original meaning by checking semantic similarity (SBERT cosine similarity) at every step and only accepts a simplification if: (1) the CEFR level hits the,target and (2) similarity to the original is above a threshold that you control (default $\ge 0.8$).

\medskip
\textbf{Prompt 2.3} I want you to update this code after the summary step, I want the program to simplify the remaining original sentence close to the target cefr\_level. After that, would you mind arranging all the JSON objects in the output file according to dataset\_id alphabetically (for instance, 01-b1 comes after 01-a1, 02-a2 comes after 01-b1).
\end{tcolorbox}

\begin{tcolorbox}[fixbox,enhanced,title={Instruction used to fix the code},breakable=false]
<Operations:> Upload the Program 1~(Baseline) file into GPT-4o

\textbf{Fix 2.1} Fix the program so that the output file contain only the text\_id column and simplified\_sentence

\medskip
\textbf{Fix 2.2} Because in the output file, there are still some JSON objects missing. So I ask GPT-4o: Can you try to update the code above so that it can simplify the original sentence of each JSON object to the target level?

\medskip
\textbf{Fix 2.3} At the end of the program, would you mind adding some code that checks the number of JSON objects in the input file with the number of the JSON objects in the output file. If they are equal, you don't need to check. If not, you need to check what the dataset\_id is missing and then simplify that original sentence belonging to that dataset\_id until all the dataset\_ids are in the output file?
\end{tcolorbox}

\subsubsection{Code Generation}
\label{sssec:mrs-rule-code}
Prompts 2.1, 2.2, and 2.3 are the three main prompts that we used to generate the code for the MRS-rule method step by step. When using Prompt 2.2 to instruct GPT-4o for further simplification by jointly checking CEFR level and semantic similarity, it suggests the following rules and is used in a sophisticated candidate generation pipeline~(\S\ref{sssec:gen-rules}).

\begin{itemize}
  \item \texttt{replace\_words}: substitute complex words with simpler synonyms (e.g., ``utilize'' $\rightarrow$ ``use'', ``approximately'' $\rightarrow$ ``about'').
  \item \texttt{simplify\_numbers\_units}: standardize numerical expressions and units (e.g., remove separators, normalize ``metres/meters'').
  \item \texttt{strip\_relative\_clauses}: remove non-essential subordinate clauses (e.g., clauses beginning with \emph{which/that/who/where/when} or discourse markers like \emph{however/although}) to reduce syntactic complexity.
  \item \texttt{keep\_shortest\_clause}: select the simplest clause from a multi-clause sentence by choosing the shortest well-formed segment.
  \item \texttt{trim\_to\_limit}: shorten the text to a step-dependent word budget while preserving a grammatical ending.
  \item \texttt{sentence\_split}: break long sentences into shorter, more readable parts at punctuation boundaries, then simplify each part.
\end{itemize}

More importantly, it also smartly suggested sacrificing semantic preservation for higher CEFR-level accuracy, demonstrating improved performance over prompting baseline~(Table \ref{tab:metrics}).

\subsubsection{Workflow}
\label{sssec:rule-workflow}
\begin{figure}[t]
\centering
\resizebox{0.95\columnwidth}{!}{%
\begin{tikzpicture}[
  font=\small,
  node distance=6mm and 10mm,
  >=Latex,
  box/.style={draw, rounded corners, align=center, inner sep=2.5pt, minimum height=6.5mm, text width=38mm},
  decision/.style={draw, diamond, aspect=2.1, align=center, inner sep=1.5pt},
  arrow/.style={->, line width=0.4pt}
]

\node[box] (step1) {Step 1: Try to produce a target-level simplification for each \texttt{text\_id}};
\node[decision, below=of step1] (miss1) {Any items missing?};

\node[box, right=16mm of miss1] (retry) {Step 2: Relax thresholds and retry on the missing set only\\[-2pt]\footnotesize (lower similarity floor, allow stronger edits)};
\node[decision, below=of retry] (miss2) {Still missing?};
\node[box, below=of miss2] (fallback) {Step 3: Fill remaining with nearest-level or conservative option};

\node[box, below=10mm of miss1] (sort) {Step 4: Sort outputs (e.g., CEFR order within group)};
\node[box, below=of sort] (write) {Step 5: Save to the output file \{\texttt{text\_id}, \texttt{simplified}\}};

\draw[arrow] (step1) -- (miss1);

\draw[arrow] (miss1) -- node[midway, left]{No} (sort);
\draw[arrow] (miss1) -- node[midway, above]{Yes} (retry);

\draw[arrow] (retry) -- (miss2);
\draw[arrow] (miss2) -- node[midway, right]{Yes} (fallback);
\draw[arrow] (fallback.west) |- (sort.east);
\draw[arrow] (miss2.west) |- node[pos=0.2, above]{No} (sort.east);

\draw[arrow] (sort) -- (write);

\end{tikzpicture}%
}
\caption{Workflow of \textbf{MRS-Rule} (Run 2)}
\vspace{-2em}
\label{fig:rule-workflow}
\end{figure}
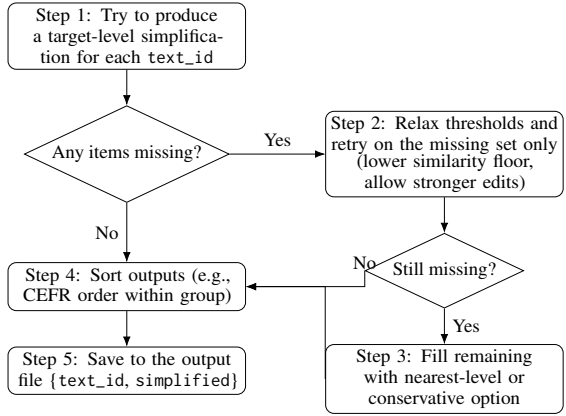

Figure~\ref{fig:rule-workflow} shows the workflow of MRS-Rule, which includes iterative retries with dynamic conditions such as similarity floor, maximum editing steps to reach the best-effort CEFR-levels. 
\paragraph{Reconciliation Retries} In each retry, the system first generates multiple candidate sentences from the original text. Then, the best candidate is selected using cosine similarity and the predicted CEFR level. This candidate becomes the seed for the next round, based on the assumption that easier sentences can be further simplified toward the target CEFR level. Candidates are created using one or more rules (details in \S\ref{sssec:gen-rules}). After each round, all candidates are scored for meaning preservation (cosine similarity) and difficulty (CEFR level). The best-scoring candidate is carried forward as the seed for the next round. If it still does not reach the target level, additional rule-based refinements are applied (\S\ref{sssec:mrs-rule-code}). Subsequent retries follow the same process, but use more relaxed thresholds. The CEFR level is validated by majority vote from three ModernBERT classifiers. Sentences that remain unsimplified go through further retries with gradually looser similarity thresholds and larger edit budgets. Finally, the system picks the candidate closest to the target level, reorders the text IDs, and outputs the results.
If any sentences are still not simplified, the system slightly lowers the similarity threshold (to 0.88), increases the maximum edit steps (to 8), and reprocesses only the remaining sentences—up to six rounds. All hyperparameters for our program are summarized in the Appendix Table~\ref{tab:hyper-mrs}.
\paragraph{Nearest-level Fill} The simplification will continue for multiple rounds of the above simplification rules until all sentences are simplified to the target level or a retry cap is reached. For sentences that did not be simplified to the target level, we will use the \textbf{nearest-level fill}, selecting the candidate whose predicted CEFR level is closest to the target while keeping the original meaning, before reorganizing and saving the final output.

\subsection{MRS-Joint: Rule-based + Prompting}
\label{ssec:pg3}
Building upon the Baseline model~(\S\ref{sec:pg1}), we combine LLM prompts~\cite{barayan-etal-2025-analysing} and rule-based multi-round simplifications with automatic verification steps~(\S\ref{ssec:pg2}). As shown in the workflow~(\S\ref{figure:mrs-joint-workflow}), the LLM generates simplified sentences only in the first step. After each retry, the system selects the best candidate on the basis of cosine similarity and predicted CEFR level. This loop continues until the predicted CEFR level matches the target level. In each new retry, the system lowers the cosine similarity threshold (allowing more meaning change) and increases the maximum number of simplification steps. This process ensures that the final sentence fits the target proficiency level while preserving the original meaning.

\subsubsection{Code Generation}
\label{sssec:pg2-code}
For the MRS-Joint program~(\S\ref{ssec:detail-program3}), we use Prompt 3.1 to integrate the LLM prompt from Baseline~(Program 1) into the MRS-Rule~(Program 2) by uploading the two program files first and then prompting. Then we use Prompt 3.2 to generate the code for over-generation-then-rank. The program, generated when we combined those two files, worked well, so there was nothing to fix.

\begin{tcolorbox}[promptbox,enhanced,title={Prompt for Generating MRS-Joint},breakable=false]
<Operations:> Upload the Program 2(MRS-Rule) and the Program 1(Baseline) to ChatGPT.

\textbf{Prompt 3.1} Update this file(the file contains the program 2). Before simplifying the sentence, the program uses the naive prompt to generate one candidate. Other candidates will be generated based on the built-in rules.

\medskip
\textbf{Prompt 3.2} After generating many candidates, the program selects the best candidate based on the cosine similarity and predicted level. If that best candidate does not meet the target level, the program continue generates more candidates based on that best candidate.

\end{tcolorbox}

\subsubsection{Workflow}
\begin{figure}[!h]
\centering
\begin{adjustbox}{max width=\linewidth, max height=.5\textheight}
\begin{tikzpicture}[
  >=Latex,
  node distance=6mm and 14mm,
  font=\footnotesize,
  state/.style={draw, rounded corners, align=left, fill=white, inner sep=2.2mm, minimum height=7mm, text width=5.6cm},
  decision/.style={diamond, aspect=2.0, draw, align=center, inner sep=0.8mm, text width=3.2cm},
  io/.style={draw, rounded corners, align=left, fill=gray!8, inner sep=2.0mm, minimum height=7mm, text width=5.6cm},
  lab/.style={midway, font=\footnotesize, inner sep=1pt, fill=white, draw=none}
]

\node[io]                             (in)      {\textbf{Inputs}\\ \texttt{original}, \texttt{target\_cefr} (+ optional \texttt{reference})};

\node[state,   below=7mm of in]       (prompt)  {\textbf{State 1 — Prompt-based}\\ \emph{build\_prompt} $\to$ \emph{call\_llm} $\to$ \emph{clean\_response}\\ Checks: \emph{cosine\_sim} $\ge$ floor; \emph{predict\_cefr} $\le$ target\\ (if ref) \emph{cosine\_sim(reference,cand)} high};

\node[decision, below=5mm of prompt]  (d1)      {Accept LLM candidate?};

\node[state,   below=8mm of d1]       (rules)   {\textbf{State 2 — Built-in Rules}\\ Generate: \emph{basic\_candidates()} (\emph{replace\_words}, \emph{simplify\_numbers\_units}, \emph{strip\_relative\_clauses}, \emph{keep\_shortest\_clause}, \emph{trim\_to\_limit}, \emph{sentence\_split})\\ Score: \emph{predict\_cefr} (3-head vote), \emph{cosine\_sim} to source (+ref), Select: best valid (hit bonus if CEFR $\le$ target)};

\node[decision, below=5mm of rules]   (d2)      {Accept rules candidate?};

\node[io, right=30mm of d1]           (out)     {\textbf{Output}\\ Accepted simplified sentence};

\node[io, below=6mm of out, fill=gray!12] (fallback) {\textbf{If neither succeeds after retries}\\ \emph{nearest\_level\_fill()} (closest CEFR while meaning preserved)\\ else trimmed fallback of original};

\draw[->] (in) -- (prompt);
\draw[->] (prompt) -- (d1);

\draw[->] (d1) to[out=0, in=180] node[lab, above] {Yes} (out.west);

\draw[->] (d1) -- node[lab, left] {No} (rules);

\draw[->] (rules) -- (d2);

\draw[->] (d2) to[out=0, in=180] node[lab, above] {Yes} (out.west);

\draw[->] (d2) to[out=-5, in=180] node[lab, above] {No} (fallback.west);

\end{tikzpicture}
\end{adjustbox}
\caption{Workflow of \textbf{MRS-Joint}}
\vspace*{-10pt}
\label{figure:mrs-joint-workflow}
\end{figure}
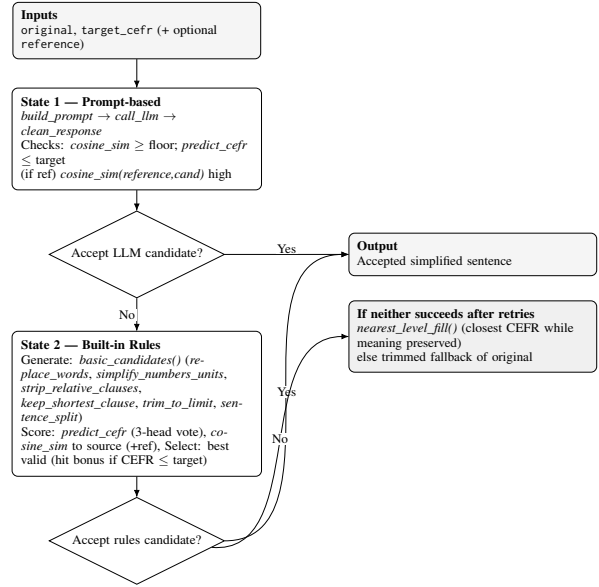

Figure~\ref{figure:mrs-joint-workflow} illustrates the \textbf{MRS-Joint} generated program by combining LLM prompting~(\S\ref{ssec:baseline}) and multi-round rule-based simplification~(\S\ref{ssec:pg2}). The generated program simply prompts the LLM in the first round, and then all subsequent rounds are purely rule-based, as described in \S\ref{sssec:rule-workflow}.

\section{Result}
\label{sec:result}
\begin{table}[!hbtp]
\centering
\small
\setlength{\tabcolsep}{6pt}
\begin{tabular}{c|c|c|c}
\toprule
\textbf{Model} & \textbf{RMSE} & \textbf{MB-Orig} & \textbf{MB-Ref} \\
\midrule
\multicolumn{4}{c}{Trial} \\ \hline
Baseline (Run 1)& 0.8944 & 0.8453 & 0.7958 \\
MRS-Rule (Run 2) & 0.8515 & 0.7961 & 0.7967 \\
MRS-Joint & 0.4472 & 0.8023 & 0.7574 \\ \hline
\multicolumn{4}{c}{Test} \\ \hline
Baseline (Run 1)& 0.755 & 0.855 & 0.849 \\
MRS-Rule (Run 2)& 0.714 & 0.865 & 0.701 \\
MRS-Joint & 0.552 & 0.866 & 0.837 \\
\bottomrule
\end{tabular}
\caption{CEFR accuracy (RMSE) and meaning preservation on trial and test datasets.}
\label{tab:metrics}
\end{table}

\begin{figure}[t]
\centering
\begin{minipage}{0.40\textwidth}
\centering
\begin{tikzpicture}
  \begin{axis}[
      width=\linewidth,
      height=0.60\linewidth,
      xlabel={Retry},
      ylabel={\#tgt-lvl reached},
      xmin=1, xmax=6,
      ymin=150, ymax=200,
      xtick={1,2,3,4,5,6},
      ymajorgrids, xmajorgrids,
      grid style={dashed,gray!30},
      legend style={
        at={(0.5,1)},
        anchor=south,
        legend columns=-1,
        font=\small,
        draw=black,
        rounded corners=2pt,
        inner sep=2pt,
      },
      legend image post style={scale=0.7},
      tick label style={font=\small},
      label style={font=\small},
      title style={font=\large},
      thick,
      clip=false,                 
      enlarge y limits={abs=6pt}, 
  ]

    \addplot[
      color=blue,
      mark=square*,
      mark size=2.5pt,
      line width=1.2pt
    ] coordinates {(1,157) (2,159) (3,165) (4,171) (5,175) (6,175)};
    \addlegendentry{MRS-Rule}

    \addplot[
      color=red,
      mark=*,
      mark size=2.5pt,
      line width=1.2pt
    ] coordinates {(1,173) (2,179) (3,185) (4,189) (5,191) (6,192)};
    \addlegendentry{MRS-Joint}

  \end{axis}
\end{tikzpicture}
\vspace{-2em}
\caption{\# Simplified sentences that reach the target level across retries for MRS-Rule vs.\ MRS-Joint.}
\label{fig:mrs_halfpage}
\end{minipage}
\end{figure}
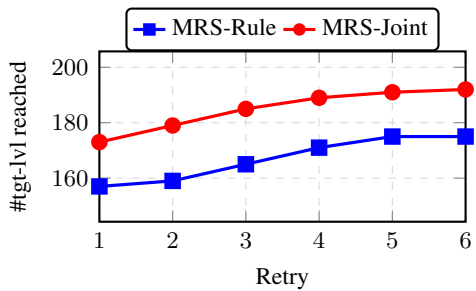

Table~\ref{tab:metrics} summarizes the performance of our models in both trial and test datasets. Baseline~(\S\ref{sec:pg1}) and MRS-Rule~(\S\ref{ssec:pg2}) are the two models corresponding to the two runs of our submission in the final evaluation period. After the evaluation, we found that simply merging two methods into MRS-Joint~(\S\ref{ssec:pg3}) is more efficient, which is the most accurate model to match the target CEFR level (the best \textbf{RMSE}) while still maintaining the meaning. The Prompt-only baseline model~(\S\ref{sec:pg1}) preserves the original meaning best (highest \textbf{MeaningBERT-Orig}) but shows the weakest control of CEFR level (highest \textbf{RMSE}). Furthermore, comparing MRS-Joint with the Baseline, the difference mainly exists in the multi-round rule-based simplification. It shows that our multi-round rules significantly improve the performance with a little sacrifice on the meaning preservation. Figure~\ref{fig:mrs_halfpage} further shows that single round simplification performs poorly, while multi-round simplification could increasingly simplify more sentences to the target CEFR level. Furthermore, MRS-Joint, starting the simplification from LLM-simplified candidates, could boost the performance of multi-round simplification. 


\section{Qualitative Analysis}
\subsection{Overall Findings}
\begin{table}[t]
\centering
\small
\setlength{\tabcolsep}{5pt}
\renewcommand{\arraystretch}{1.05}
\begin{tabular}{lrrrrrr}
\toprule
\textbf{Target}\,$\downarrow$\ /\ \textbf{Pred}\,$\rightarrow$ & \textbf{A2} & \textbf{B1} & \textbf{B2} \\
\midrule
\textbf{A2} & 66 & 32 & 2 \\
\textbf{B1} & 20 & 79 & 1\\
\textbf{B2} & 0 & 0 & 0 \\
\bottomrule
\end{tabular}
\caption{Confusion matrix on the test data}
\label{tab:confmat}
\vspace{-2em}
\end{table}
As shown in Table~\ref{tab:confmat}, the program excels in simplifying complex sentences C1-C2 to the B1 level. Of 100 source sentences, 79 were successfully simplified to B1, with only 20 dropping further to A2 and 1 rising to B2. For sentences targeted at the A2 level, the results were mixed: only 66 reached the intended A2 level, while 32 overshot B1 and 2 even remained at B2.

Therefore, we recognize that simplifying the input of high-complexity C1–C2 to lower CEFR levels is inherently more challenging. The program is more prone to “overshooting,” producing text that remains more complex than the intended target. In other words, the lower the target CEFR level, the higher the likelihood of program's not meeting the constraints of that level.

\subsection{Case Study}
To understand the behavior of the model beyond the overall accuracy scores, we performed a \textbf{qualitative error analysis} on three representative examples misclassified by the CEFR predictor. These examples
illustrate three different types of misclassification.
\paragraph{Case 1 – Overshoot (A2 $\rightarrow$ B1)~(\S\ref{ssec:case1})} 
The model simplified vocabulary and shortened clauses but kept abstract ideas along with a relative clause typical of the B1 syntax. 
The CEFR predictor therefore rated the output B1, which is one level higher than the target level, showing that preserving key ideas may force more complex structures than the intended level.
\paragraph{Case 2 - Lexical Imitation~(\S\ref{ssec:case2})}
Although shortened from the source text, the output kept formal phrases like “a large number of bridge accidents... of the bridge itself"” instead of simpler A2 wording such as “Many accidents happen while bridges are being built.” The CEFR model therefore rated it B1, showing that better simplification requires lexical adaptation, not just shorter text.
\paragraph{Case 3 – Under-generation (B1 $\rightarrow$ A2)~(\S\ref{ssec:case3})} 
The system produced only a fragment, dropping the telescope's purpose and the planetary-defense discussion. 
With much of the conceptual content missing, the predictor judged the text A2 despite technical terms. 
This highlights that incomplete outputs can seem easier to cheat the CEFR predictor as the intended CEFR level.

These examples reveal three failure modes --overshoot, and undergeneration -- demonstrating that successful CEFR simplification requires not only simpler words but also balanced control of meaning, style, and completeness.

\section{Conclusion}
We found that a larger gap between the CEFR level of the original and target sentences~(CEFR-Gap) increases the likelihood of simplification failure. Based on this finding, we proposed two multi-round simplification methods generated by GPT-4o: MRS-Rule, which applies rule-based simplification, and MRS-Joint, which combines rules with prompting. Extensive experiments and case studies show that MRS-Joint outperforms both the prompting baseline and MRS-Rule, confirming the effectiveness of multi-round simplification and the feasibility of text simplifyer via code generation.
\section*{Limitation}
We note a few limitations of our work. The models we used are closed-source models such as using GPT-4o for code generation while using GPT-4o-mini for API, which are not explicitly finetuned in the text simplification datasets by us. Our work is also limited to one dataset and one language (English), and two types of GPT-4o generated model. Furthermore, focusing on coding generation, we could also extend the study to self-evolve algorithm discovery~\cite{novikov2025alphaevolve} and compare it with other prompts and more coding agents. Besides those, we believe explicitly involving curriculum-based domain knowledge in a structured multi-round simplification will be promising methods in the era of artificial intelligence.

\section*{Lay Summary}
This project aims to make complex English sentences easier to understand, especially for language learners. Our team participated in the TSAR 2025 competition, in which the goal was to rewrite sentences to match specific levels of English proficiency, such as beginner (A1) or intermediate (B1), based on the \textbf{Common European Framework of Reference (CEFR)}.
The insight of our team was that the greater the difference between the original difficulty of a sentence and the target level (called the “CEFR Gap”), the harder it is to simplify the sentence successfully. For example, turning a very advanced sentence (C1) into a basic one (A2) is much more difficult than making small adjustments. This inspired us to develop a \textbf{multistep approach} for simplification.

Our team created two systems, and the code is generated with AI with our instructions:

\textbf{MRS-Rule}: Uses rules to gradually simplify text in multiple rounds (e.g., replace difficult words, break long sentences).

\textbf{MRS-Joint}: Combines a model (GPT-4o-mini) to generate an initial simplified text, and then refines it through multiple rule-based steps.

Both systems repeatedly check whether the new sentence meets the desired CEFR level and still retains the original meaning. If not, they retry those sentences with adjustments. This multi-round process continues until the system either succeeds or picks the closest acceptable version.

In testing, the MRS-Joint method performs best. It reaches the target reading level more often than the baseline approach, although sometimes at the cost of slightly reducing the original meaning. Still, it shows strong overall results: it handles complex sentences better and produced more accurate simplifications.
Our team also analyzed the errors. Sometimes, the program “oversimplified” or retained too many complex words. Other times, it shortened the sentence too much and left out important information. These findings will help improve future systems.

In short, this work shows that a multi-step process can make content more accessible to learners while maintaining its original intent.

\bibliography{custom}






\appendix

\section{Details for 3 Generated Program}
\label{sec:detail-code}
In general, all three programs are generated by GPT-4o model, which covers the following python libraries and models. 
\paragraph{Libraries} In the AI generated code of MRS-Rule and MRS-Joint,  the following Python libraries are used: \texttt{argparse}, \texttt{os}, \texttt{sys}, \texttt{json}, \texttt{re}, \texttt{math}, \texttt{pathlib}, \texttt{collections}, \texttt{typing}, \texttt{numpy}, \texttt{requests}. 
Besides those regular pythong libraries, we noticed that libraries like \texttt{transformers} (Hugging Face), \texttt{SentenceTransformers}, \texttt{NumPy} are used for natural language processing and machine learning parts. 
\paragraph{Models} CEFR level is predicted by three ModernBERT from huggingface, ModernBERT-base-doc\_en-Cefr, ModernBERT-base-doc\_sent\_en-Cefr, ModernBERT-base-reference\_AllLang2-Cefr; while semantic similarity is using the library of~\texttt{sentence transformers/all-MiniLM-L6-v2}), and LLM for generating the code is GPT-4o\footnote{\url{https://chatgpt.com/?model=gpt-4o}, accessible at 09/23/2025}. The LLM API used for text simplification is \texttt{gpt-4o-mini}.

\subsection{Program 1: Baseline \Naive~Prompt}
\label{ssec:detail-program1}
The generated Program 1 is in the file of "First\_Version\_Sentence\_Simplification.py" in the code repo. It is built with \textbf{OpenAI's Chat Completions API}. The script is lightweight and designed for \textbf{large-scale, reproducible simplification} runs, while maintaining a clean JSONL output compatible with downstream CEFR or readability evaluations. This baseline Program 1 is used as Run 1 in our submission and is also used in our \textbf{CEFR-Gap} analysis.

\subsection{Program 2: MRS-Rule}
\label{ssec:detail-program2}
The generated Program 2 is in the file "Second\_Version\_Sentence\_Simplification.py". 
Specifically, ChatGPT suggests useful rules to generate candidate implication with a basic candidate \texttt{base \_candidates()}. Generate multiple simplified variants of an input sentence using lightweight rule-based transformations without relying on an LLM. The details of the code are shown in the code listing ~\ref{lst:code}. The corresponding hyperparameters used in the code are summarized in Table~\ref{tab:hyper-mrs}. 

\begin{table}[!hbtp]
\centering
\small
\setlength{\tabcolsep}{6pt}

\begin{tabular}{@{}ll@{}}
\toprule
\textbf{Parameter} & \textbf{Value} \\
\midrule
\texttt{similarity\_floor} & 0.88  \\
\texttt{max\_steps} & 8 \\
\texttt{max\_retries} & 6 \\
\texttt{floor\_step} & 0.03  \\
\texttt{steps\_step} & 6  \\
\texttt{sim\_floor (internal)} & 0.88 $\downarrow$\\
\texttt{w\_hit} & 10  \\
\texttt{w\_ref} & 2.5  \\
\texttt{w\_orig} & 0.5 \\
\texttt{llm\_timeout} & 60~seconds \\
\texttt{use\_llm} & \texttt{true} \\
\texttt{sim\_threshold} & 0.72--0.75  \\
\bottomrule
\end{tabular}
\caption{Hyperparameters in MRS-Rule and MRS-Joint}
\label{tab:hyper-mrs}
\end{table}

\subsubsection{Generated Code to Apply Rules}
\label{sssec:gen-rules}

\begin{lstlisting}[language=Python, caption={Rule-based Simplification to Generate Candidates}, label={lst:code}]
base = text.strip()
lim = max(8, 28 - 2*step_idx)
cands = [
    replace_words(base),
    simplify_numbers_units(base),
    keep_shortest_clause(base),
    strip_relative_clauses(base),
    trim_to_limit(replace_words(base), lim),
    trim_to_limit(simplify_numbers_units(base), lim),
    trim_to_limit(strip_relative_clauses(base), lim),
    trim_to_limit(keep_shortest_clause(replace_words(base)), lim),
]
for s in sentence_split(strip_relative_clauses(base)):
    cands.append(trim_to_limit(replace_words(s), max(10, lim-4)))
\end{lstlisting}

\subsection{Program 3: MRS-Joint}
\label{ssec:detail-program3}

Program 3 is in the file "Third\_Version\_Sentence\_Simplification.py". It simply combines the previous rule-based program~\ref{ssec:detail-program2} with LLM-prompting-based program in~\ref{ssec:detail-program1}. This program shares the same rules used in~\ref{sssec:gen-rules}, and uses the same set of multi-round hyper parameters as Table~\ref{tab:hyper-mrs}.

\section{Case Studies}
\label{sec:appendix}
To understand the model’s behaviour beyond overall accuracy scores, we conducted a
\textbf{qualitative error analysis} on three representative C1--C2 source texts that were
simplified by the system but misclassified by the CEFR predictor. These examples
illustrate three distinct types of misclassification.

\subsection{Case 1 -- Overshoot: A2 Target $\rightarrow$ Predicted B1}
\label{ssec:case1}

\textbf{Source (C1--C2)}\\
\emph{I am a member of the Campaign to Stop Killer Robots \ldots\ We need to be very cautious indeed about what we ask machines to do.}

\textbf{System Output (intended A2)}\\
\emph{I am part of the Campaign to Stop Killer Robots. Forget the scary robots from movies. Right now, military machines like drones and robots are already used to kill with little help from people. The next step is robots that can decide who to kill by themselves. I believe this is very wrong and dangerous for people. We need to be very careful about what we let machines do.}

\textbf{Analysis}\\
The model successfully shortened the complex original sentence that had many clauses and replaced high-level terms
(\emph{autonomous ‘murderbots’}, \emph{completely unethical}) with simpler words.
However, it retained \textbf{abstract concepts} (\emph{robots that can decide who to kill})
and an \textbf{embedded relative clause}, both typical of B1 syntax.
The CEFR predictor therefore judged the output \textbf{one level higher than the A2 target}.\\
\emph{Insight:} Even when sentences are lexically simplified, preserving key ideas may
require structures that naturally push the text above the intended level.

\subsection{Case 2 -- Lexical Imitation: A2 Target $\rightarrow$ Predicted B1}
\label{ssec:case2}

\textbf{Source (C1--C2)}\\
\emph{A large number of bridge accidents occur during the construction of the bridge itself \ldots\ this can be deadly for the workers on it at the time.}

\textbf{System Output (intended A2)}\\
\emph{A large number of bridge accidents occur during the construction of the bridge itself. These accidents are often due to.}

\textbf{Analysis}\\
Although truncated, the output \textbf{copied formal phrases from the source}
(\emph{a large number of bridge accidents occur}, \emph{during the construction of the bridge itself}).
It failed to rephrase these into simpler, high-frequency A2 forms such as
\emph{Many accidents with bridges happen while they are being built.}
The CEFR model therefore still rated it \textbf{B1}, despite the missing ending.\\
\emph{Insight:} True simplification requires \textbf{lexical adaptation}, not only shortening.
Retaining formal academic expressions—even in a shorter text—can maintain a higher perceived level.

\subsection{Case 3 -- Under-generation: B1 Target $\rightarrow$ Predicted A2}
\label{ssec:case3}

\textbf{Source (C1--C2)}\\
\emph{Whether NASA can find the remaining middle-sized NEOs depends on getting the money to build NEOCam \ldots\ the PHAs.}

\textbf{System Output (intended B1)}\\
\emph{NASA's ability to find the remaining middle-sized near-Earth objects (NEOs) depends on getting funding to build NEOCam, a 0.5-meter space telescope that.}

\textbf{Analysis}\\
The system produced only a \textbf{partial sentence}, omitting the telescope’s function
and the entire discussion of planetary defence.
With the \textbf{conceptual load drastically reduced}, the CEFR predictor assigned an
\textbf{A2 level}, even though the fragment still contains technical terms (\emph{NEOs}, \emph{NEOCam}).\\
\emph{Insight:} Incomplete outputs can appear easier than intended, causing the CEFR
assessment to \textbf{underestimate} the level. Quality checks for completeness are
essential alongside automatic scoring.

\end{document}